\documentclass[conference]{IEEEtran}
\IEEEoverridecommandlockouts
\usepackage{cite}
\usepackage{amsmath,amssymb,amsfonts}
\usepackage{algorithmic}
\usepackage{graphicx}
\usepackage{textcomp}
\usepackage{xcolor}
\usepackage{enumitem}
\usepackage{booktabs}
\usepackage{multirow}
\usepackage{gensymb}
\usepackage{flushend}
\usepackage{siunitx} 
\usepackage[a4paper, total={184mm,239mm}]{geometry}
\def\BibTeX{{\rm B\kern-.05em{\sc i\kern-.025em b}\kern-.08em
    T\kern-.1667em\lower.7ex\hbox{E}\kern-.125emX}}
\begin{document}

\title{Causal-Guided Dimension Reduction for Efficient Pareto Optimization}

\author{Dinithi Jayasuriya, Divake Kumar, Sureshkumar Senthilkumar, Devashri Naik, Nastaran Darabi and Amit Ranjan Trivedi \\
University of Illinois at Chicago, IL, USA, email: \{dkasth2, dkumar33, ssenth7, dnaik6, ndarab2, amitrt\}@uic.edu
\vspace{-1.75em}}

\maketitle

\begin{abstract}
Multi-objective optimization of analog circuits is hindered by high-dimensional parameter spaces, strong feedback couplings, and expensive transistor-level simulations. Evolutionary algorithms such as Non-dominated Sorting Genetic Algorithm II (NSGA-II) are widely used but treat all parameters equally, thereby wasting effort on variables with little impact on performance, which limits their scalability. We introduce \textbf{CaDRO}, a causal-guided dimensionality reduction framework that embeds causal discovery into the optimization pipeline. CaDRO builds a quantitative causal map through a hybrid observational-interventional process, ranking parameters by their causal effect on the objectives. Low-impact parameters are fixed to values from high-quality solutions, while critical drivers remain active in the search. The reduced design space enables focused evolutionary optimization without modifying the underlying algorithm. Across amplifiers, regulators, and RF circuits, CaDRO converges up to 10$\times$ faster than NSGA-II while preserving or improving Pareto quality. For instance, on the Folded-Cascode Amplifier, hypervolume improves from 0.56 to 0.94, and on the LDO regulator from 0.65 to 0.81, with large gains in non-dominated solutions.
\end{abstract}

\vspace{5pt}
\begin{IEEEkeywords}
Analog Circuit Design, Multi-Objective Optimization, Causal Inference, Design Automation
\end{IEEEkeywords}

\section{Introduction}

Analog circuit design remains one of the most intricate problems in EDA: designers must simultaneously satisfy conflicting objectives such as gain, bandwidth, power, noise, and area across dozens of interdependent device and bias parameters. These parameters interact nonlinearly through feedback loops, parasitics, and biasing networks, making the search space difficult to navigate. Brute-force or black-box optimization is prohibitively expensive for complex circuits and does not reveal which parameters truly drive performance, limiting critical design insights and their reuse across topologies, domains, and application constraints \cite{TleloCuautle2010_EA_AnalogSizing,rashid2023performance}.

To reduce simulation cost, prior work can be grouped into four main directions: \textit{First}, surrogate and Bayesian optimization methods for analog circuit sizing, with recent efforts tackling high-dimensional BO via subspace or truncated sampling \cite{Gu2024_tssbo}. \textit{Second}, ML-based mapping and heuristic methods predict parameters directly from specifications or guide search via learned domain knowledge (e.g., LEDRO’s use of LLMs for design space reduction) \cite{Kochar2024_LEDRO}. \textit{Third}, RL and variation-aware optimization address real-world PVT variation and multi-task scenarios; RobustAnalog employs multi-task RL with pruning \cite{Shi2022_RobustAnalog}, while ROSE-Opt combines BO+RL with domain knowledge for robust optimization \cite{Cao2024_RoseOpt}. \textit{Fourth}, dimensionality or search-space reduction techniques shrink parameter count or ranges via statistical methods, subspace selection, or sparse regression, often guided by feature importance or gradient approximations (e.g., Sparse Regression \& Error Margining \cite{Alawieh2016_EfficientSparse}, LinEasyBO \cite{Zhang2021_LinEasyBO}, LEDRO \cite{Kochar2024_LEDRO}).

\begin{figure}[t!]
    \centering
    \includegraphics[width=\linewidth]{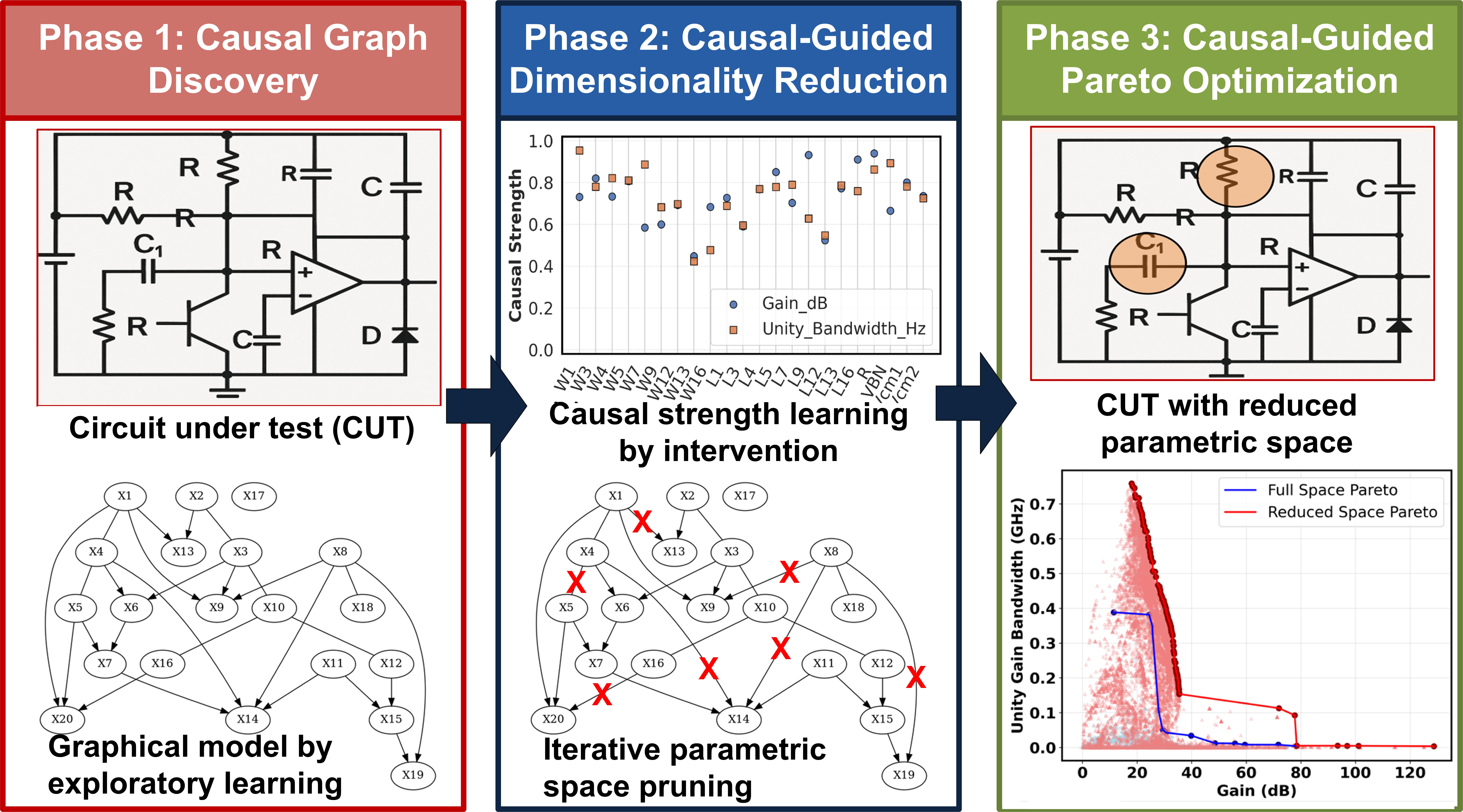} \vspace{-20pt}
    \caption{\textbf{CaDRO framework}: Phase~1 learns the causal graph of the circuit under test (CUT). Phase~2 prunes weak parameters via causal strength estimation. Phase~3 performs Pareto optimization on the reduced space, achieving near full-space performance with lower complexity.}\vspace{-5pt}
    \label{fig:fig1}
\end{figure}

While prior works have improved efficiency, they share a key limitation: parameter importance is typically inferred from correlation or sensitivity. In analog circuits, where dense interconnections and feedback loops are pervasive, such measures are often misleading. Variables may appear predictive through indirect coupling while having little direct effect on performance. As a result, optimizers waste simulations on parameters that do not meaningfully affect the Pareto front. 

We address this gap by introducing \textit{causal reasoning} as a systematic basis for dimensionality reduction (Fig. 1). Causal analysis provides a principled way to separate true design drivers from spurious associations. Unlike deep learning models that capture correlations without interpretability, causal reasoning makes dependencies explicit and actionable for design. 

With this rationale, this paper introduces CaDRO: a causal-guided dimensionality reduction framework that embeds causal discovery into multi-objective evolutionary search for analog circuit design. We integrate CaDRO into the Non-dominated Sorting Genetic Algorithm II (NSGA-II) to assess its impact on multi-objective analog circuit optimization. In CaDRO, an initial exploratory run generates a diverse dataset, from which we construct a causal map using a hybrid observational–interventional approach. Each parameter is assigned a causal strength, allowing us to separate critical design drivers from low-impact variables. The latter are fixed to values from high-performing solutions. The evolutionary algorithm then operates only on the reduced set, yielding order-of-magnitude simulation savings while preserving Pareto quality across analog and RF benchmarks. Across diverse analog circuits including amplifiers, regulators, and RF circuits, CaDRO converges up to 10$\times$ faster than baseline NSGA-II while preserving or improving Pareto front quality. Beyond efficiency, causal maps reveal which parameters truly govern circuit behavior, making the approach scalable and interpretable.

\section{Background and Related Work}


\subsection{Causal Discovery}

Causal discovery uncovers directed cause–effect relationships among variables, moving beyond correlations. A causal graph $G=(V,E)$ represents parameters and performance objectives as nodes $V$, with directed edges $E$ denoting direct influences. Unlike correlation graphs, which capture statistical associations, causal graphs encode asymmetric, interventionally testable dependencies~\cite{scholkopf2022causality , klein2022correlation , spirtes2016causal}. Common methods include:
\begin{itemize}[leftmargin=*]
  \item \textit{Constraint-based algorithms} such as PC~\cite{spirtes2000causation} and FCI~\cite{spirtes2000causation,zhang2008causal} infer causal directions from conditional independencies.  
  \item \textit{Score-based searches} (e.g., GES~\cite{chickering2002optimal}) optimize a scoring criterion such as Bayesian information criterion (BIC).  
  \item \textit{Functional causal models} exploit asymmetries in data-generating processes: LiNGAM~\cite{shimizu2006lingam} assumes linear non-Gaussian models, while NOTEARS~\cite{zheng2018notears} casts discovery as continuous optimization.  
\end{itemize}

Observational data alone are often insufficient due to confounding: two parameters may appear correlated not because one causes the other, but because both are driven by hidden factors. To resolve this, \textit{interventions}, i.e, actively perturbing one parameter while holding others fixed, are used to confirm edges and quantify causal effect sizes~\cite{pearl2009causality}. Distinguishing true design drivers from spurious correlations is critical in analog circuits, where dense feedback loops and shared current paths often create misleading associations. Causal maps that quantify each parameter’s effect and confidence provide a principled basis for dimensionality reduction before costly multi-objective optimization. \textit{Yet}, prior works have not closed this loop. E.g., Jiao et al.~\cite{jiao2016analog} extracted parameter-performance dependencies from simulations but did not integrate causal knowledge into optimization. Other approaches, such as surrogate-assisted Bayesian optimization (MACE~\cite{lyu2018batch}), reinforcement learning under PVT variation (RobustAnalog~\cite{shi2022robustanalog}), deterministic group-based search~\cite{xu2024deterministic}, and hybrid evolutionary–surrogate schemes~\cite{liu2009analog} improve efficiency and Pareto quality, but still treat all variables as equally important; thereby facing scalability constraints as parametric space expands.

\subsection{Non-Dominated Sorting Genetic Algorithm II (NSGA-II)}

The Non-Dominated Sorting Genetic Algorithm II (NSGA-II)~\cite{deb2002fast} is a widely used evolutionary algorithms for multi-objective optimization. Like other genetic algorithms, it maintains a population of candidate solutions and evolves them through selection, crossover, and mutation, with three  features:

\begin{itemize}[leftmargin=*]
  \item \textit{Fast non-dominated sorting}: Ranks solutions by Pareto dominance for scalable multi-objective selection.  
  \item \textit{Crowding distance}: Preserves diversity by favouring solutions in less crowded regions of the Pareto front.  
  \item \textit{Elitism}: Ensures the best non-dominated solutions are retained across generations.  
\end{itemize}

NSGA-II has been extended to other variants, including NSGA-III~\cite{deb2013evolutionary} for many-objective problems and MOEA/D~\cite{zhang2007moea} for decomposition-based optimization. In analog circuit design, NSGA-II has been widely applied to trade off gain, bandwidth, noise, and power~\cite{salgado2010applications,rashid2023performance}, but its efficiency deteriorates as the number of variables grows, since all parameters are perturbed and evolved regardless of their true influence. This motivates dimensionality reduction techniques of CaDRO that shrink the search space while preserving solution quality.

\section{CaDRO: Causal-Guided Dimensionality Reduction for Multi-Objective Optimization}

CaDRO is a three-phase pipeline that accelerates multi-objective optimization by reducing the effective dimensionality of the design space before expensive evolutionary search. Its core idea is to discover and exploit the causal structure linking design parameters to performance metrics by analyzing true cause–effect relationships. By quantifying and ranking these relationships, CaDRO directs computation to a small subset of influential parameters while safely fixing those with negligible impact. This selective focus yields substantial simulation savings without compromising Pareto quality. The pipeline consists of three phases: (i) Causal Discovery and Strength Analysis, (ii) Causal-Based Dimensionality Reduction, and (iii) Focused Multi-Objective Optimization. We discuss each below:

\subsection{\underline{Phase 1}: Causal Discovery and Strength Analysis}
CaDRO begins by constructing a high-confidence map of cause–effect relationships within the circuit. To this end, we generate a comprehensive dataset through an exploratory run of NSGA-II that evolves a population of candidate solutions via selection, crossover, and mutation, while employing non-dominated sorting and crowding distance to balance convergence and diversity \cite{deb2002fast}. \textit{Notably}, the goal here is not to find optimal solutions but to leverage NSGA-II’s exploration capability. By sampling a broad range of parameter combinations, the algorithm produces a dataset containing good, bad, and mediocre designs with solution diversity that is essential for capturing statistical relationships across the design space. In our framework, causal discovery is performed using 8k simulations, which also serve as the initial population for subsequent optimization.

With this dataset, we begin the observational analysis. In complex analog circuits, parameter–performance relationships are often highly non-linear, so a single analytical method risks missing key dependencies. To address this, we employ an ensemble approach combining Pearson correlation for linear trends, Random Forest models for non-linear effects, and mutual information for general dependencies. An initial confidence score is synthesized from this evidence: it starts with a base score from the statistical significance ($p$-value) of the Pearson correlation, then is incrementally boosted if the link is also supported by other methods, such as high feature importance from tree-based models. This yields a robust preliminary score reflecting consensus across the ensemble.

To distinguish true causality from spurious correlations, we perform \textit{interventional refinement}, an active learning process with targeted experiments on the most uncertain links \cite{varici2024interventional , hagele2023bacadi}. Each intervention runs two small sets of simulations: in the first, a single input parameter is fixed to a ``low'' value (e.g., its 25th percentile), and in the second to a ``high'' value (e.g., its 75th percentile), while other parameters vary. A statistical test compares the output distributions of the two groups: if a significant difference is observed, the causal link is confirmed and its confidence score upgraded; if not, the link is classified as spurious with confidence set to zero. The causal strength (effect size) is then quantified as the normalized difference in the output mean between the ``high'' and ``low'' groups, providing a direct measure of the parameter’s influence. In this way, we experimentally confirm or reject each causal hypothesis. After several interventions, the final output is a weighted directed graph, with each edge annotated by causal strength and a refined confidence score based on combined observational and interventional evidence\cite{li2023causal}.

\begin{figure*}[t!]
    \centering
    \includegraphics[width=0.85\linewidth]{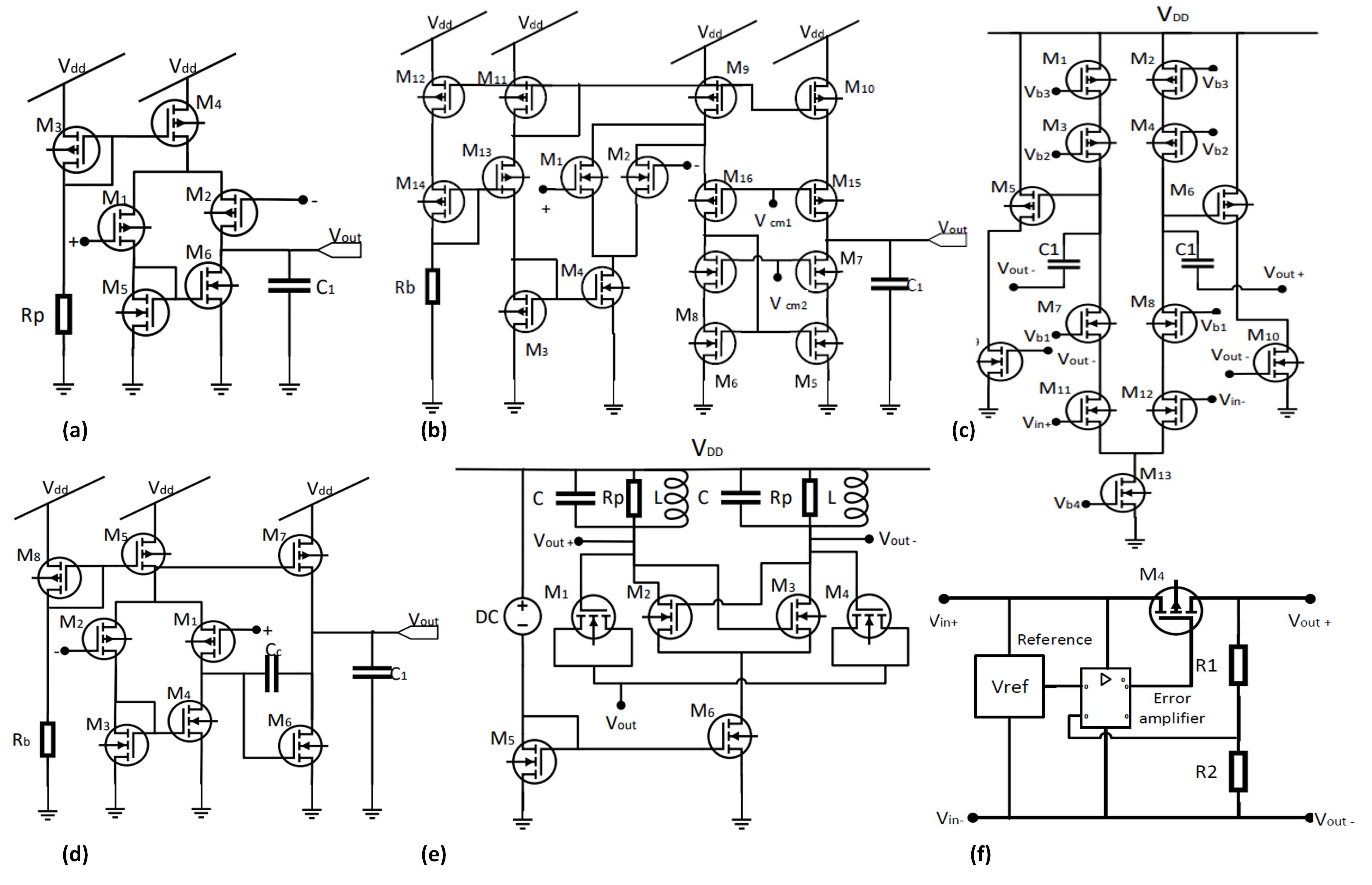} \vspace{-15pt}
\caption{\textbf{Schematics of benchmark circuits used to evaluate CaDRO:} (a) Active-Load Differential Amplifier, (b) Folded-Cascode Amplifier, (c) Two-Stage Voltage Amplifier, (d) Two-Stage Operational Transconductance Amplifier, (e) Voltage-Controlled Oscillator, and (f) Low-Dropout Regulator.}\vspace{-10pt}
    \label{fig:tsva_obj}
\end{figure*}

\begin{figure*}[t!]
    \centering
    \includegraphics[width=0.85\linewidth]{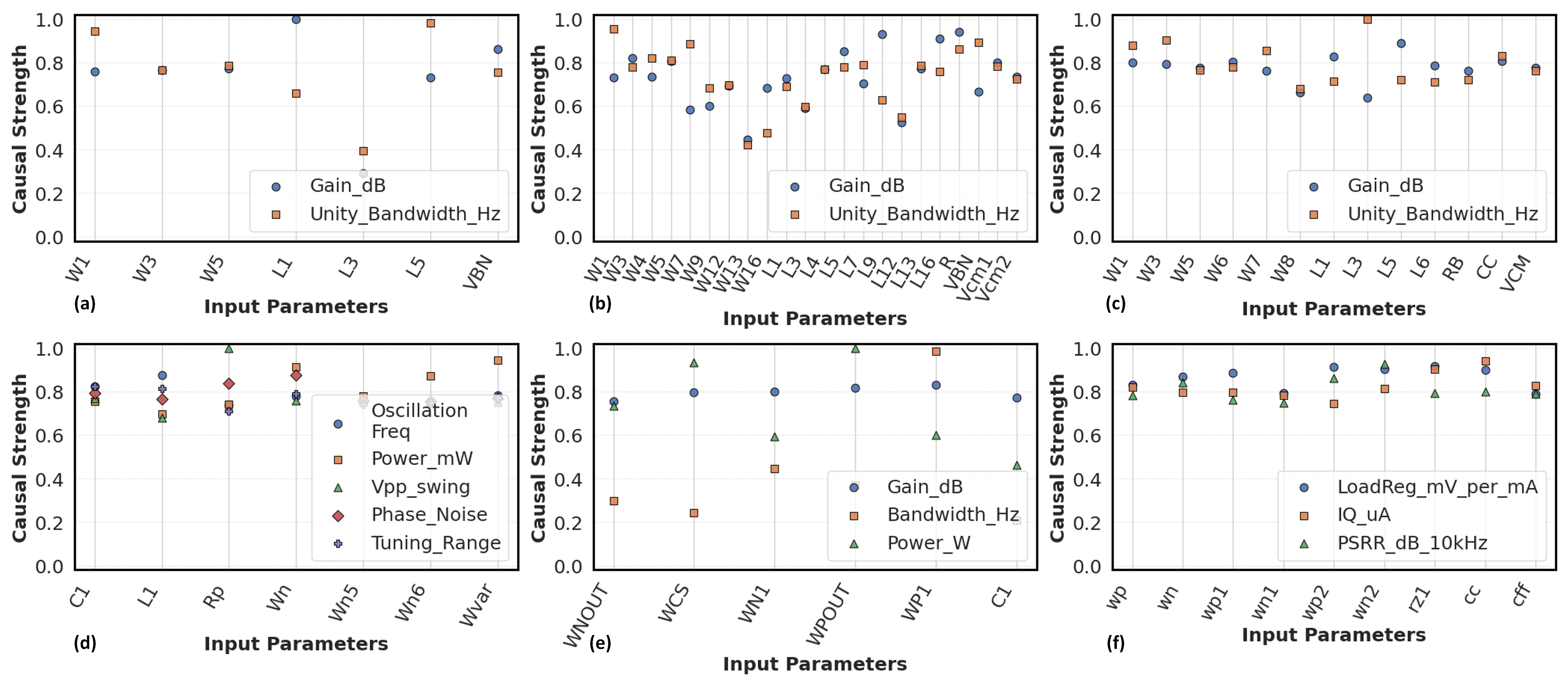} \vspace{-10pt}
\caption{\textbf{Causal strength of input parameters across benchmark circuits:} 
Each subplot shows normalized causal strengths from CaDRO's observational–interventional analysis. 
(a) ALDA, (b) FCA, (c) TSOTA, (d) VCO, (e) TSVA, and (f) LDO. 
Markers denote circuit objectives (e.g., Gain, UGBW, Power, PSRR). The profiles highlight dominant drivers and low-impact parameters that can be pruned safely.}\vspace{-5pt}
    \label{fig:causal_strengths}
\end{figure*}

\subsection{\underline{Phase 2}: Causal-Based Dimensionality Reduction}

With the validated causal map from Phase~1, we obtain a reliable guide to which parameters truly drive circuit performance. Phase~2 uses this knowledge to simplify the optimization problem by focusing computational effort on influential parameters while filtering those with negligible impact. \textit{First}, we compute an overall \textit{importance score} for each parameter. For a given input $P_i$, the score $S(P_i)$ is defined as
\[
S(P_i) = \sum_j \big| E(P_i, O_j) \times C(P_i, O_j) \big|,
\]
where $E(P_i, O_j)$ is the causal effect size of $P_i$ on objective $O_j$, and $C(P_i, O_j)$ is the corresponding confidence score. This formulation holistically captures total influence: the product weights effects by their reliability, the absolute value treats negative and positive effects equally, and the summation across objectives identifies parameters with broad influence as more critical than those with narrow but strong effects. Based on these scores, design parameters are partitioned into an \textit{Active set}, containing the most influential parameters to be optimized, and a \textit{Pruned set}, containing low-importance parameters removed from active search. Our framework supports both fixed (\textit{top-$k$}) and adaptive pruning strategies, with the cutoff determined by the distribution of importance scores. Crucially, pruned parameters are not discarded or set arbitrarily, which could shift the design into suboptimal regions. Instead, they are fixed to values drawn from high-performing, non-dominated solutions obtained in Phase~1, anchoring the search in a proven region and providing a strong foundation for final optimization.

\subsection{\underline{Phase 3}: Focused Multi-Objective Optimization}

The final phase performs multi-objective optimization on the reduced problem. NSGA-II is initialized to operate exclusively in the low-dimensional search space defined by the active parameters from Phase~2. All evolutionary mechanisms, population management, crossover and mutation, and selection, are thus focused solely on this critical subset.

For each candidate solution, only the active parameters are generated. To enable simulation, the framework reconstructs a full parameter vector by combining these active values with the fixed values of pruned parameters. This ensures evaluation in a valid, high-performance region while avoiding exploration of inconsequential dimensions. By shrinking the search space, the optimizer sidesteps flat, low-impact regions, leading to an exponential reduction in volume and much faster convergence. This focused approach reduces both the number of simulations and the total time needed to obtain high-quality Pareto solutions.

\section{Results and Evaluation}

\subsection{Benchmark Circuits and Design Objectives}

We evaluate CaDRO on six benchmark circuits spanning amplifiers, regulators, and RF blocks. The Active-Load Differential Amplifier (ALDA) has seven parameters and aims to maximize low-frequency gain and UGBW while minimizing power (Fig. 2a). The Two-Stage OTA (TSOTA) includes 13 parameters, targeting high DC gain and UGBW with low power (Fig. 2d). The Folded-Cascode Amplifier (FCA) has 22 parameters and seeks high gain and bandwidth at low power (Fig. 2b). The Low-Dropout (LDO) Regulator, with nine parameters, minimizes load regulation and quiescent current while maximizing PSRR at 10\,kHz (Fig. 2f). The Voltage-Controlled Oscillator (VCO) optimizes seven parameters of a cross-coupled LC tank to maximize oscillation frequency and minimize power (Fig. 2e). Finally, the Two-Stage Voltage Amplifier (TSVA) has six parameters and targets high gain and UGBW with low power (Fig. 2c). Mainly ablating the utility of causal reasoning for automated analog EDA, our experiments address two crucial questions: (i) how does the degree of causal-guided pruning trade off computational efficiency and solution quality, and (ii) how do the final optimization results from the pruned search space compare to traditional optimization? 

\begin{figure*}[t!]
    \centering
    \includegraphics[width=0.95\linewidth]{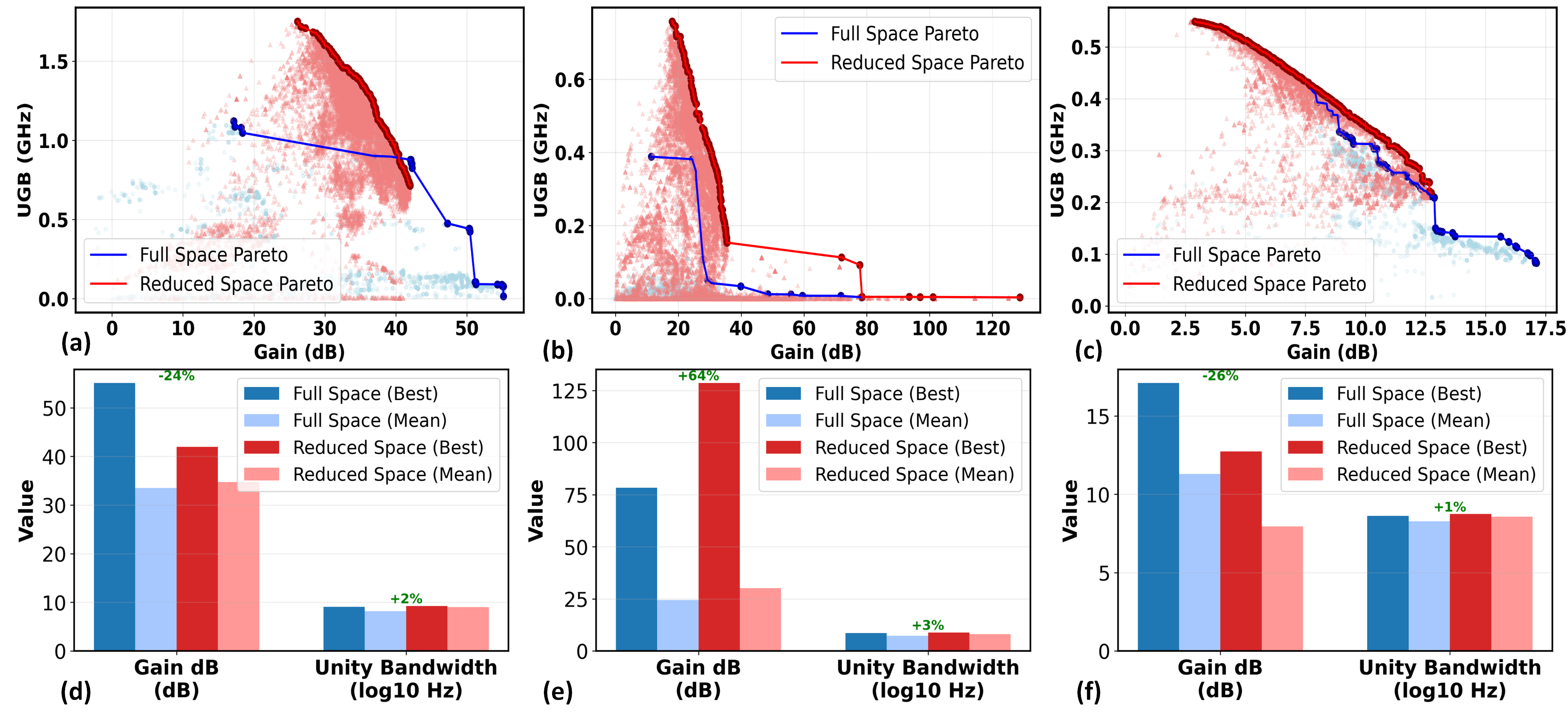} 
\caption{\textbf{Amplifier benchmarks under CaDRO vs. full-space NSGA-II:} 
(a–c) Pareto fronts for the Two-Stage OTA (TSOTA), Folded-Cascode Amplifier (FCA), and Active-Load Differential Amplifier (ALDA), comparing reduced-space CaDRO (red) against full-space NSGA-II (blue). CaDRO consistently produces outward-shifted and denser Pareto fronts. 
(d–f) Best and mean values of Gain and Unity-Gain Bandwidth from the corresponding fronts. In TSOTA (d), CaDRO sacrifices some gain (–24\%) but slightly improves UGBW (+2\%). In FCA (e), CaDRO achieves a large gain improvement (+64\%) together with higher UGBW (+3\%). In ALDA (f), gain decreases (–26\%) while UGBW is preserved (+1\%). 
These plots show that causal pruning redistributes search effort: in circuits where causal drivers strongly govern both objectives (FCA), fronts expand dramatically, while in others (TSOTA, ALDA), fronts densify and trade-offs shift.}\vspace{-5pt}

\label{fig:2D}
\end{figure*}

\subsection{Discovered Causal Strengths}
In the causal discovery phase, we quantify the influence of each design parameter on performance objectives. We visualize the causal discovery results as per-parameter scatter panels as shown in Fig. 3. In each panel, the x-axis lists the input parameters, the y-axis reports the normalized causal strength in [0, 1], and different objectives are distinguished by legend-coded markers. A higher point on the y-axis means that parameter has a stronger direct influence on that objective; points near 0 indicate little to no influence. In FCA and TSOTA, most parameters show strong causal links, so aggressive pruning degrades performance; in FCA nearly all transistor dimensions affect both Gain and UGBW. In contrast, ALDA exhibits more selective dependencies: L1 and L5 (differential pair and load) dominate Gain, while the tail current source (L3) has weak influence and is prunable. The other benchmarks also match design intuition. In the LDO, the pass device width ($W_{p2}$) strongly impacts both LoadReg and PSRR, consistent with its role in output regulation. In the VCO, the tank inductor ($L_1$) directly determines oscillation frequency. In TSVA, the input pair width ($W_{P1}$) drives first-stage transconductance and bandwidth, while the output device ($W_{POUT}$) primarily sets power and moderately increases bandwidth. These causal maps confirm known dependencies while identifying low-impact parameters.

\subsection{Impact of Causal Pruning on Pareto Trade-Offs}

Fig.~\ref{fig:2D} compares reduced-space CaDRO (red) against full-space NSGA-II (blue) for three amplifier circuits: TSOTA, FCA, and ALDA. The top row shows Pareto front overlays in Gain vs. Unity-Gain Bandwidth (UGBW), while the bottom row reports best and mean values with percentage changes. These views show not only where CaDRO improves or shifts the fronts, but also \textit{how causal pruning redistributes search effort across competing objectives}. In the Two-Stage OTA (TSOTA, Fig.~\ref{fig:2D}a,d), the reduced-space front covers the trade-off curve more densely, reflecting more efficient exploration. The best-case gain drops (–24\%), while mean and best UGBW values improve slightly (+2\%). This indicates that CaDRO favours balanced designs with more reliable bandwidth at the cost of extreme gain, often preferable in practical analog design where stability margins matter.  

For the Folded-Cascode Amplifier (FCA, Fig.~\ref{fig:2D}b,e), CaDRO achieves the most dramatic improvement: the Pareto front shifts outward along both axes, hypervolume increases from 0.56 to 0.94, and best-case gain improves substantially (+64\%) while UGBW also rises (+3\%). The denser frontier shows that pruning low-impact parameters prevents wasted evaluations and concentrates sampling where true trade-offs exist. In the Active-Load Differential Amplifier (ALDA, Fig.~\ref{fig:2D}c,f), CaDRO maintains full frontier coverage but with a noticeable trade-off shift: gain decreases (–26\%) while UGBW is essentially preserved (+1\%). The result is a denser but lower-gain frontier, showing that in smaller circuits with fewer strong causal drivers, pruning can emphasize bandwidth consistency at the expense of peak gain. Importantly, mean performance remains competitive, indicating that CaDRO avoids collapsing the search space.    

\begin{figure*}[t!]
    \centering
    \includegraphics[width=0.95\linewidth,]{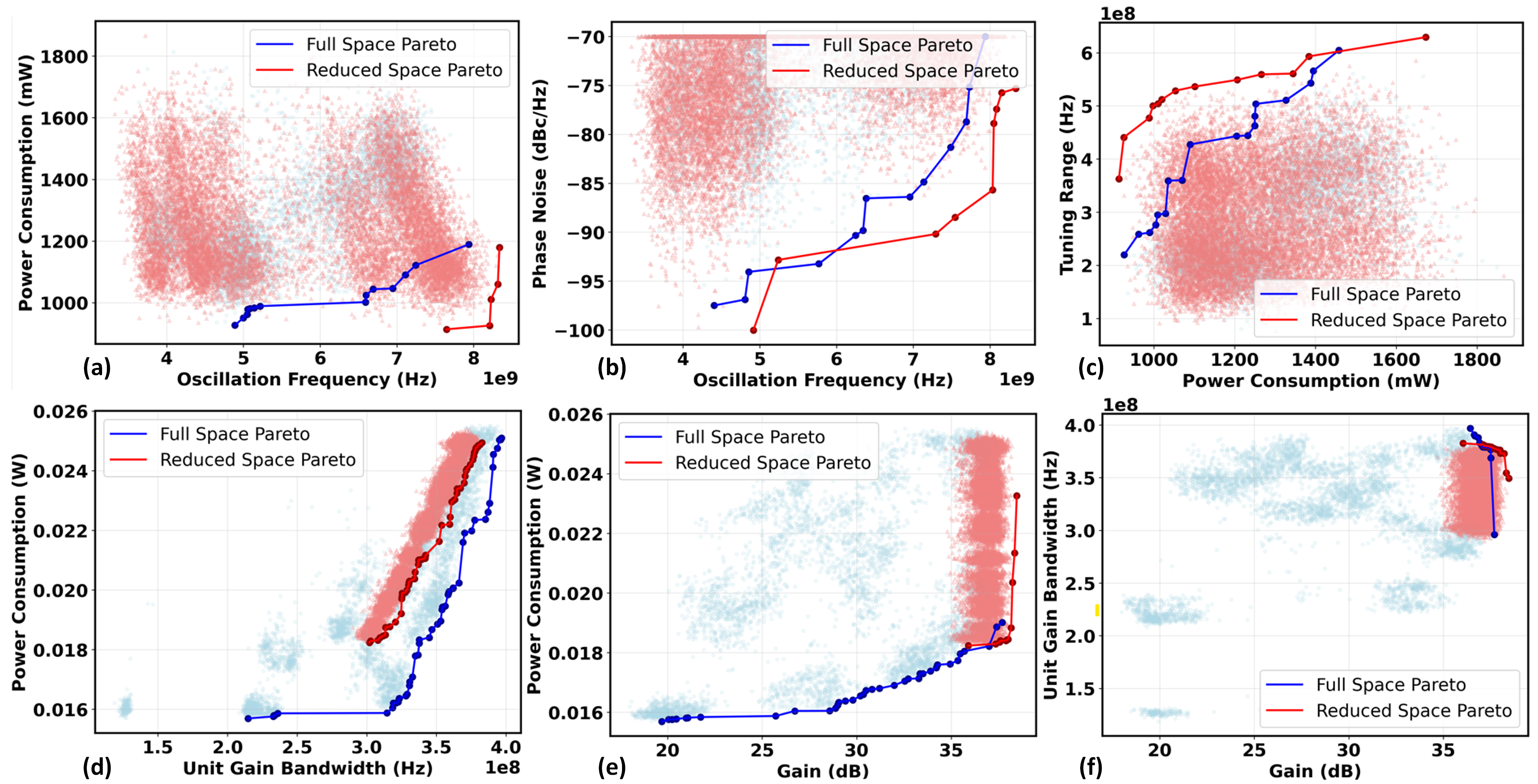} 
\caption{\textbf{Oscillator and Regulator benchmarks under CaDRO vs. full-space NSGA-II:} 
(a–c) Voltage-Controlled Oscillator (VCO) and (d–f) Two-Stage Voltage Amplifier (TSVA). 
Reduced-space CaDRO (red) achieves fronts nearly identical to full-space NSGA-II (blue) but with far fewer evaluations. 
In the VCO, CaDRO maintains trade-offs across oscillation frequency, power, phase noise, and tuning range, while avoiding wasted exploration of weakly causal parameters. 
In the TSVA, fronts in gain, bandwidth, and power consumption closely track the baseline yet are reached with an order-of-magnitude fewer simulations.
}
    \label{fig:causal_refinement}
\end{figure*}

Fig. 5 extends this comparison to the Voltage-Controlled Oscillator (VCO) and Two-Stage Voltage Amplifier (TSVA). In the VCO (Fig. 5a–c), CaDRO preserves the full set of trade-offs in oscillation frequency, power, phase noise, and tuning range. The close overlap of the red and blue fronts shows that pruning weakly causal parameters does not distort achievable performance, while efficiency improves substantially; fronts are reached with less than half the evaluations (1160 $\rightarrow$ 478). 
For the TSVA (Fig. 5d–f), the impact is even clearer. Causal analysis revealed that only the input pair and compensation capacitor drive behavior. Once the remaining parameters are fixed, CaDRO reproduces the same gain, bandwidth, and power trade-offs as the baseline, but with an order-of-magnitude fewer evaluations (49 $\rightarrow$ 4). 

\begin{table*}[t!]
\centering
\setlength{\tabcolsep}{6pt}
\textbf{Table I: Quantitative comparison of CaDRO and full-space NSGA-II across amplifier and regulator benchmarks.}\vspace{2pt}
\label{tab:quantitative_results}
\begin{tabular}{@{}ll
                S S S S S S S
                r@{}}
\toprule
\textbf{Circuit} & \textbf{Method (Sim. count(\%))} &
\textbf{Hypervol. (↑)} & \textbf{GD (↓)} & \textbf{IGD (↓)} &
\textbf{Additive $\varepsilon$ (↓)} & \textbf{Spacing S (↓)} &
\textbf{Coverage} & \textbf{Cardinality} \\
\midrule
\multirow{2}{*}{FCA (2D)} 
  & Reduced Space (100\%) & 0.944265 & 0.444278 & 0.409681 & 0.717150 & 1.345029 & 1.000000 & 48 \\
  & Reduced Space (80\%)  & 0.34     & 0.03     & 0.66     & 0.98     & 0.87     & 0.2      & 8 \\
  & Full Space (NSGA-II)  & 0.56     & 0.142053 & 0.328009 & 0.627524 & 1.321571 & 0.020833 & 5 \\
\midrule
\multirow{2}{*}{ALDA (2D)}
  & Reduced Space(100\%)  & 0.307027 & 0.018291 & 0.089093 & 0.564061 & 0.713425 & 0.764706 & 410 \\
  & Reduced Space (80\%)  & 0.143    & 0.043    & 0.491    & 0.752    & 0.735    & 0.09     & 176 \\
  & Full Space (NSGA-II)  & 0.288488 & 0.062454 & 0.139056 & 0.364272 & 0.940782 & 0.000000 & 85 \\
\midrule
\multirow{2}{*}{TSOTA (2D)}
  & Reduced Space (100\%) & 0.612925 & 0.055853 & 0.134614 & 0.473822 & 0.773460 & 0.258065 & 231 \\
  & Reduced Space (80\%)  & 0.240789 & 0.278701 & 0.368293 & 0.554827 & 0.967511 & 0.01     & 192 \\
  & Full Space (NSGA-II)  & 0.478969 & 0.078355 & 0.148459 & 0.261962 & 1.246551 & 0.207792 & 31 \\
\midrule
\multirow{2}{*}{LDO (3D)}
  & Reduced (100\%)       & 0.810095 & 0.061130 & 0.380744 & 0.601803 & 1.445393 & 0.609756 & 602 \\
  & Reduced (80\%)        & 0.759262 & 0.051967 & 0.490029 & 0.792599 & 1.365115 & 0.447761 & 982 \\
  & Full Space (NSGA-II)  & 0.653727 & 0.125371 & 0.567203 & 0.831415 & 1.590745 & 0.125255 & 164 \\
\bottomrule
\end{tabular}
\end{table*}

\subsection{Quantitative Benchmarking}

Table I reports a detailed comparison of CaDRO and full-space NSGA-II across amplifiers and the LDO regulator. Each row lists the circuit, method, and simulation budget, followed by  metrics: hypervolume (overall volume of the dominated region, ↑), Generational Distance (GD, average distance from obtained solutions to the true front, ↓), Inverted Generational Distance (IGD, average distance from reference Pareto points to the obtained set, ↓), additive $\varepsilon$ indicator (worst-case dominance gap, ↓), spacing $S$ (distribution uniformity, ↓), coverage (fraction of baseline solutions dominated, ↑), and cardinality (size of the non-dominated set) \cite{ishibuchi2018reference , ishibuchi2015modified , panichella2022improved}.  

In the amplifier benchmarks, CaDRO consistently improves hypervolume while lowering GD and IGD. For the Folded-Cascode Amplifier (FCA), hypervolume rises from 0.56 to 0.94, and the non-dominated set expands from 5 to 48, reflecting both better coverage and richer sampling. In the Two-Stage OTA (TSOTA), hypervolume improves from 0.48 to 0.61, while spacing shrinks (1.25 $\rightarrow$ 0.77), yielding a more uniform distribution of solutions. The Active-Load Differential Amplifier (ALDA) highlights another strength: although hypervolume increases only modestly (0.29 $\rightarrow$ 0.31), the number of non-dominated solutions grows nearly fivefold (85 $\rightarrow$ 410).  

The Low-Dropout Regulator (LDO) provides perhaps the clearest evidence of causal pruning’s effect. Hypervolume increases from 0.65 to 0.81, GD is halved (0.125 $\rightarrow$ 0.061), and coverage rises from 0.13 to 0.61, while spacing decreases relative to baseline. These quantitative gains mirror the visual expansions seen earlier: CaDRO consistently reallocates simulation effort away from low-impact parameters and into regions of genuine trade-off, producing higher-quality and more comprehensive Pareto sets at lower computational cost.  

\begin{table}
\setlength{\tabcolsep}{5pt}
\centering
\textbf{Table II: Comparison of CaDRO and full-space NSGA-II for RF circuits (VCO and TSVA) across convergence and diversity metrics.}\vspace{2pt}
\begin{tabular}{@{}llccccccccccccccc@{}}
\toprule
\textbf{Circuit} & \textbf{Method} & \textbf{GD} & \textbf{IGD} & \textbf{S} & \textbf{Delta} & \textbf{MS} & \textbf{CP}    \\
\midrule
\multirow{2}{*}{VCO} & Reduced Space & 0.11 & 0.16 & 0.02 & 0.39 & 0.75& 478   \\
& Full Space & 0.10 & 0.07 & 0.01 & 0.42 & 0.92  & 1160    \\
\midrule
\multirow{2}{*}{TSVA} & Reduced Space & 0.03 & 0.25 & 0.08 & 0.88 & 0.25  & 4    \\
& Full Space & 0.0298 & 0.11 & 0.01 & 0.99 & 0.88  & 49   \\
\bottomrule
\end{tabular}\vspace{-20pt}
\end{table}

In Table II, for the Voltage-Controlled Oscillator (VCO), CaDRO achieves nearly the same GD and IGD as the full-space baseline (0.11 vs. 0.10, 0.16 vs. 0.07) but with less than half the evaluations (1160 $\rightarrow$ 478). Spacing (S) and Delta ($\Delta$), which measure uniformity and diversity, remain close to baseline, while Maximum Spread (MS) is slightly reduced, indicating the front is preserved though sampled more compactly. Computational Cost (CP) highlights the efficiency gain directly. In the Two-Stage Voltage Amplifier (TSVA), the effect is sharper: simulations collapse from 49 to 4, yet GD and IGD remain competitive (0.03 vs. 0.0298, 0.25 vs. 0.11). Here, S and $\Delta$ increase modestly, reflecting thinner sampling, but MS and overall coverage remain intact. The near-identical Pareto surface confirms that CaDRO isolates the true performance drivers while eliminating wasted evaluations.

\section{Conclusion}
We presented CaDRO, a causal-guided dimensionality reduction framework for scalable multi-objective optimization of analog and RF circuits. By combining observational–interventional causal discovery with evolutionary search, CaDRO identifies true design drivers, prunes low-impact parameters, and anchors optimization in high-performance regions. Results across amplifiers, regulators, and oscillators show that CaDRO converges up to $10\times$ faster than NSGA-II, consistently improves or preserves Pareto quality, and yields denser, more interpretable fronts. 

\bibliographystyle{IEEEtran}
\bibliography{main}
\end{document}